# Survival tree and meld to predict long term survival in liver transplantation waiting list

Emília Matos do Nascimento [a], Basilio de Bragança Pereira [a,b,*], Samanta Teixeira Basto [b], Joaquim Ribeiro Filho [b]

[a]*Federal University of Rio de Janeiro, COPPE - Postgraduate School of Engineering, Rio de Janeiro, Brazil*

[b]*Federal University of Rio de Janeiro, School of Medicine and HUCFF - University Hospital Clementino Fraga Filho, Rio de Janeiro, Brazil*

---

**Abstract**

**Background:** Many authors have described MELD as a predictor of short-term mortality in the liver transplantation waiting list. However MELD score accuracy to predict long term mortality has not been statistically evaluated.

**Objective:** The aim of this study is to analyze the MELD score as well as other variables as a predictor of long-term mortality using a new model: the Survival Tree analysis.

**Study Design and Setting:** The variables obtained at the time of liver transplantation list enrollment and considered in this study are: sex, age, blood type, body mass index, etiology of liver disease, hepatocellular carcinoma, waiting time for transplant and MELD. Mortality on the waiting list is the outcome. Exclusion, transplantation or still in the transplantation list at the end of the study are censored data.

**Results:** The graphical representation of the survival trees showed that the most statistically significant cut off is related to MELD score at point 16.

**Conclusion:** The results are compatible with the cut off point of MELD indicated in the clinical literature.

* Corresponding author. Tel.: +55-21-25622594
E-mail address: basilio@hucff.ufrj.br (B.B. Pereira)

**What is New**

- MELD score cut off to predict long term mortality in liver transplantation waiting list was statistically evaluated for the first time.

- Survival Analysis Tree and MELD was used to predict long term mortality.

## 1. Introduction

The Model for End-Stage Liver Disease (MELD) score was described as a short term mortality index used to predict three month mortality in patients who underwent transjugular intrahepatic portosystemic shunt (TIPS) insertion [1]. It was subsequently applied to allocate liver grafts in liver transplantation list in the United States and several countries, since February 2002 [2]. Many countries use subjective local criteria or UNOS based policy to allocate liver grafts according to liver disease severity [3]. In Brazil, liver transplantation waiting list was organized according to a chronological system until June, 2006 [4].

The liver transplantation waiting list time varies significantly among various centers but usually reflect a gap between the donor liver pool and the demand for transplant [5]. The longer waiting time results in a higher mortality rate [6].

It is important to identify those patients with the worst outcome. There are several factors related to liver transplantation waiting list mortality as age, gender, blood type and disease etiology [7].

Many authors have described MELD as an independent tool related to short term mortality in the transplantation waiting list and tried to determine a threshold to assess prognosis and mortality in this setting [8,9]. However MELD score accuracy to predict long term mortality has not been statistically evaluated in the past.

The aim of this study was to analyze the MELD score as a predictor of long term mortality using a Survival Analysis Tree and to establish a MELD cut off point that better predicts this long term mortality. Cut off points of other covariates are also evaluated and their interactions with MELD is also analyzed.

The data base and methods are presented in section 2. Section 3 presents the recursive partitioning method. The results and conclusion are presented in sections 4 and 5 respectively.

## 2. Data base and methods

From November 1997 to July 2006, all patients in the liver transplantation waiting list were evaluated for inclusion in the study. Patients with incomplete data for MELD calculation were excluded and 529 were included. Data were obtained from the patient inclusion registration form and from the hospital's internal system of patient registration (Medtrack) and organized in excel for posterior analysis.

The variables obtained at the time of liver transplantation list enrollment and considered in this study are: sex, age, blood type, body mass index, etiology of liver disease, hepatocellular carcinoma, waiting time for transplant (in days) and MELD. The formula for the MELD score [1] is $3.8*\log_e(\text{bilirubin[mg/dL]}) + 11.2*\log_e(\text{INR}) + 9.6*\log_e(\text{creatinine [mg/dL]}) + 6.4*$(etiology: 0 if cholestatic or alcoholic, 1 otherwise).

From the 529 patients in the data base, 61% were male. The mean age was 51±13 years old. The most frequent etiology for liver disease was chronic hepatitis C (47%), alcoholic liver disease (17%) and cryptogenic (10%). Regarding general outcome, 36% died, and 64% are censored, from which 8% left the transplant list, 14% had been submitted to a liver transplant, and 42% are still in list.

The statistical approach used is the Survival Tree developed by Hothorn et al. [10]. The implementation was done using R [11] packages [10].

## 3. Recursive partitioning

A learning set *L* consists of m covariates $X = (X_1,..., X_m)$ of a sample space $\chi = \chi_1 \times ... \times \chi_m$ and a response Y of a sample space *Y*. Let it be a learning set *L* used to form a predictor $\varphi(x, L)$, i.e., if the input is x the answer y will be predicted by φ(x, *L*).

So, the conditional distribution $D(Y \mid X)$ of the response given covariates X depends on a function f of the covariates $D(Y \mid X) = D(Y \mid X_1,..., X_m) = D(Y \mid f(X_1,..., X_m))$, with the restriction that the partition is based on the regression relationships so that the covariate space $\chi = \bigcup_{k=1}^{r} B_k$ are partitioned in r disjoint cells $B_1,..., B_r$ .

The regression model will be fitted based on a learning sample $L_n$ composed of n independent and identically distributed observations.

Hothorn et al.[10] used regression models describing the conditional distribution of a response Y given the status of m covariates through the tree-structured recursive partitioning and formulated a generic algorithm for recursive binary partitioning for a given learning sample $L_n$ using non-negative valued case weights $w = (w_1,..., w_n)$. Each node of a tree is represented by a vector of non-zero case weights if the corresponding observations are elements of the node and zero otherwise.

The association between the response $Y$ and covariates $X_j$ , $j = 1, ... , m$ is measured by the following linear statistics

$$T_j(L_n, w) = vec\left(\sum_{i=1}^{n} w_i g_j(X_{ji}) h(Y_i, (Y_1,...,Y_n))^T\right) \in \Re^{pjq}$$

where $g_j : X_j \rightarrow \Re^{pj}$ is a non-random transformation of the covariate $X_j$ .

$h: Y \times Y^n \rightarrow \Re^q$ is the influence function that depends on the responses ($Y_1, ...,Y_n$) in a permutation symmetric way.

vec is the operator that convert a $p_j$ x q matrix into a $p_jq$ column vector by column-wise combination.

The distribution of $T_j(L_n, w)$ under the partial hypotheses $H_j^0$ depends on the joint distribution of Y and $X_j$ , which is unknown under almost all practical circumstances. This principle leads to test procedures known as permutation tests.

The majority of the algorithms for the construction of classification or regression trees algorithm follow a general rule [12]:

1) Partition the observations by univariate splits in a recursive way.

2) Fit a constant model in each cell of the resulting partition.

Hothorn et al. [10] implemented the conditional inference trees which embed recursive binary partitioning into the well defined theory of permutation tests developed by Strasser and Weber [13]. These are the steps of the algorithm [14]:

*1) Test the global null hypothesis of independence between any of the input variables and the response (which may be multivariate as well). Stop if this hypothesis cannot be rejected. Otherwise*

*select the input variable with strongest association to the response. This association is measured by a p-value corresponding to a test for the partial null hypothesis of a single input variable and the response.*

*2) Implement a binary split in the selected input variable.*

*3) Recursively repeats steps 1) and 2).*

The algorithm stops if the global null hypothesis of independence between the response Y and any of the m covariates cannot be reject at a pre-specified nominal level $\alpha$. Otherwise the association between the response and each of the m covariates is measured by test statistics or P-values that indicate the deviations from the partial hypotheses $H_j^0$.

## 4. Results

This section presents a graphical representation of the survival tree for the 529 patients in the liver transplantation waiting list using R [11] packages [10]. P-values correspond to the log-rank test.

In Figure 1 one can observe the MELD cut off at point 16. This survival tree also presents some other cut offs statistically significant.

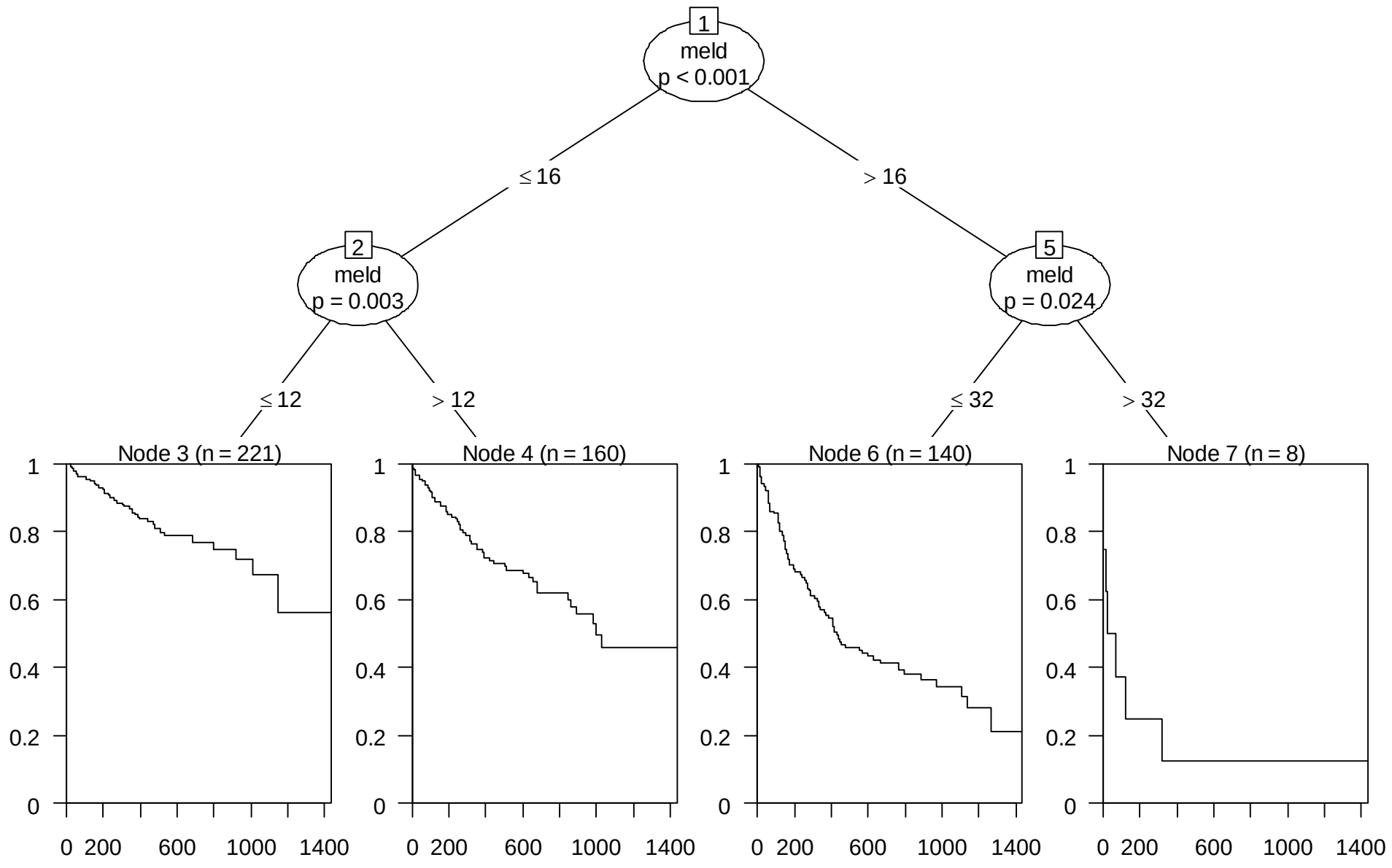

**Figure 1 – Survival Tree (MELD)**

Figure 2 shows that the first important cut off is related to MELD at point 16 and also presents the interaction with age where the cut off corresponds to 33.2 years.

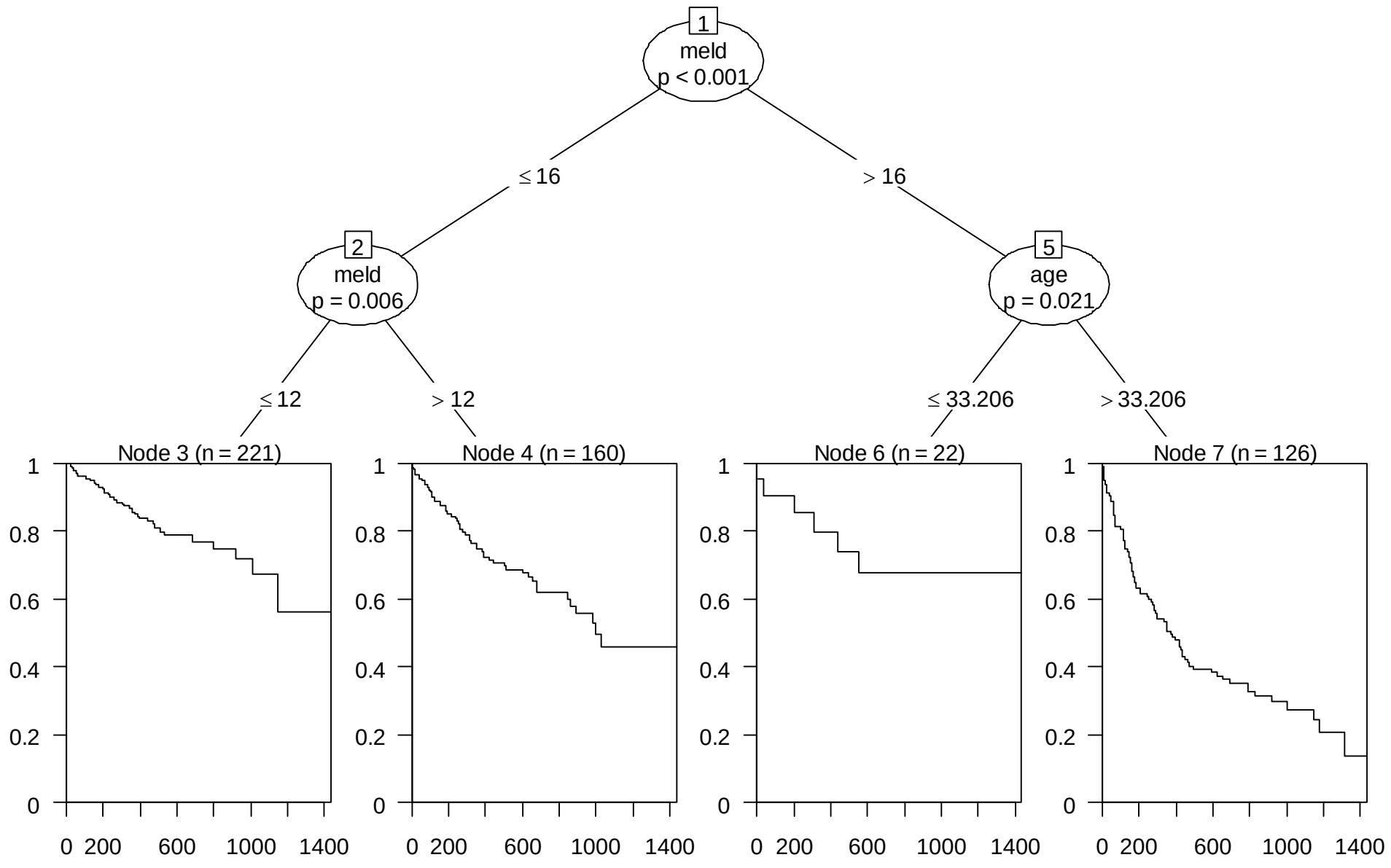

**Figure 2 – Survival Tree (Interaction between MELD and age)**

Figure 3 shows again that the principal cut off is corresponding to MELD at point 16 and also the relevant interaction with hepatocellular carcinoma (HCC).

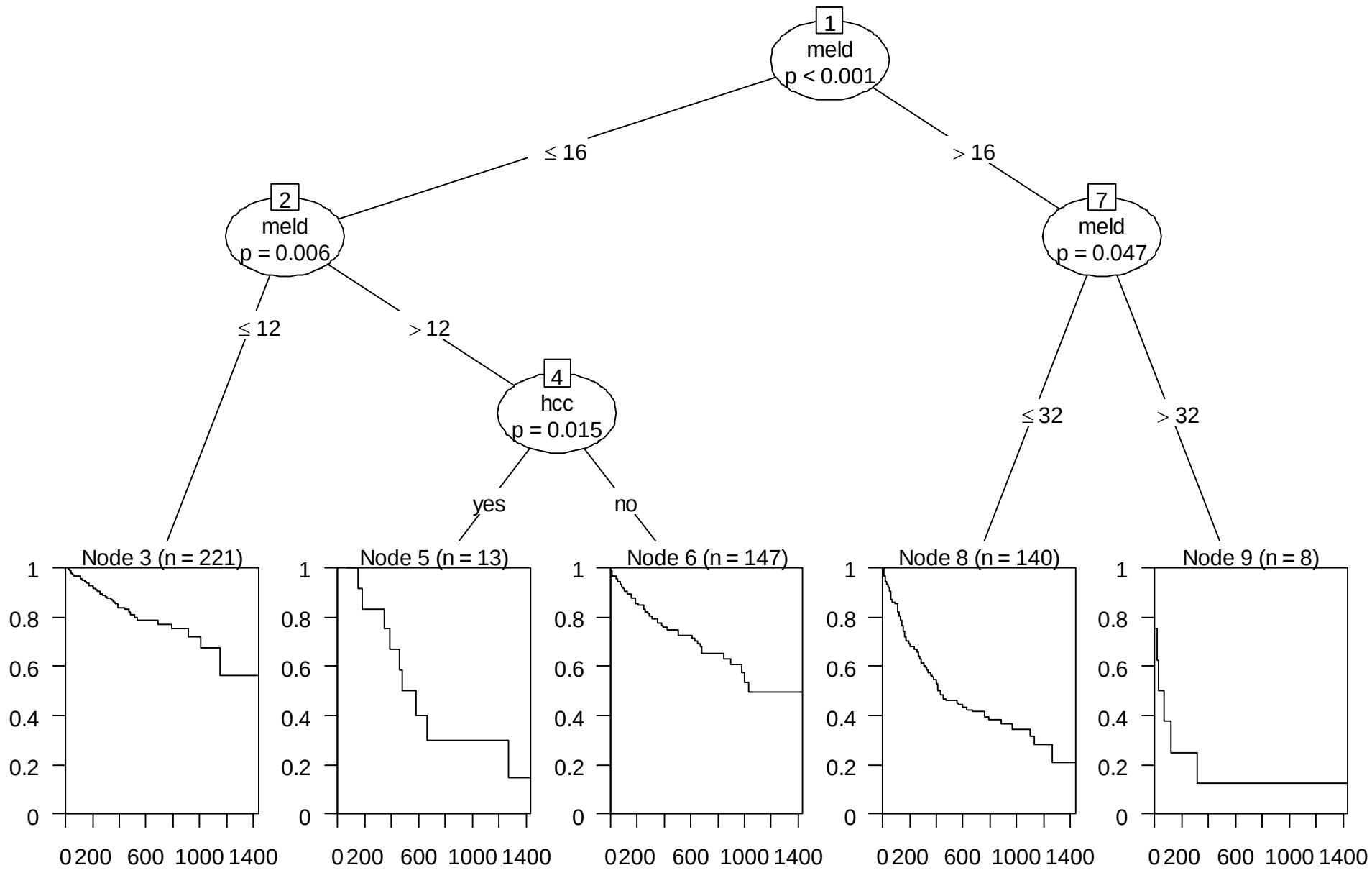

**Figure 3 – Survival Tree (Interaction between MELD and HCC)**

Finally, Figure 4 presents a decision cut off with three variables: MELD (cut off at point 16), age, and hepatocellular carcinoma (HCC). The other variables in the data base did not show any interaction with MELD.

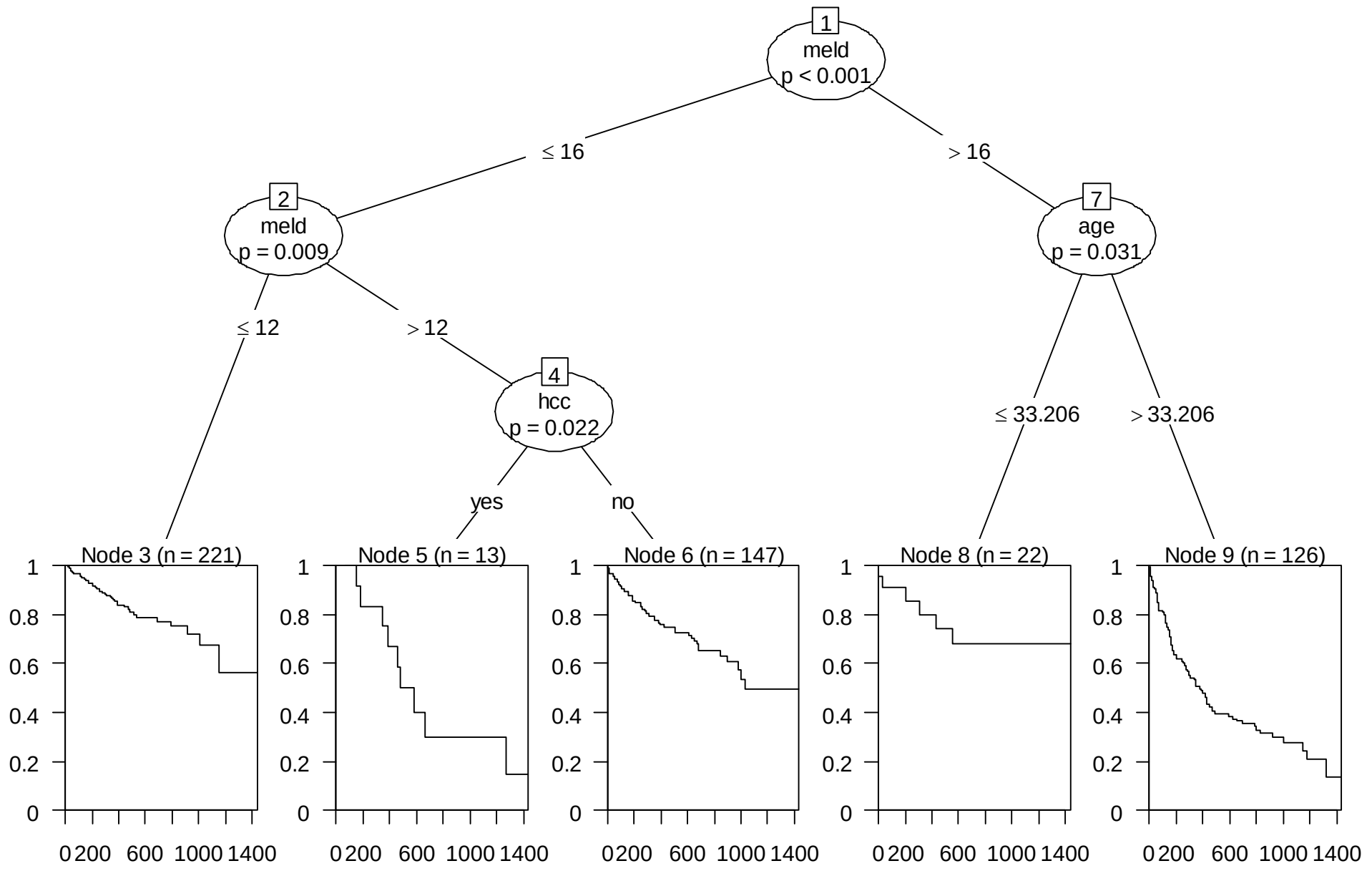

**Figure 4 – Survival Tree (Interaction between MELD, age and HCC)**

## 5. Conclusion

The optimal cut off point to MELD score in the clinical literature is around 16 or 17 [8,15]. This has been confirmed in all survival trees presented in this paper where the cut off point to MELD score based on the data is 16. Our statistical results reinforce the cut off point indicated in the clinical literature.

## Authors' contributions

Nascimento EM (M.Sc.) was responsible for the implementation of the R packages; Nascimento EM (M.Sc.) and Pereira B de B (Ph.D.) analyzed the data; Basto ST (M.D.) and Ribeiro Filho J (M.D.) collected and organized the data. All authors read and approved the final manuscript.